\documentclass{article} 
\usepackage{iclr2019_conference,times}
\usepackage{graphicx}     
\graphicspath{{./figures/}{../figures/}{../jpeg/}{./image/}}    

\usepackage{amsmath,amsfonts,bm}

\def\eqref#1{equation~\ref{#1}}

\def\1{\bm{1}}

\DeclareMathAlphabet{\mathsfit}{\encodingdefault}{\sfdefault}{m}{sl}
\SetMathAlphabet{\mathsfit}{bold}{\encodingdefault}{\sfdefault}{bx}{n}

\usepackage{hyperref}
\usepackage{url}
\usepackage{amssymb}

\newcommand{\red}[1]{\textcolor{black}{#1}}

\title{Deep Hashing using Entropy Regularised Product Quantisation Network}

\author{Jo Schlemper, Jose Caballero, Andy Aitken, Joost van Amersfoort \\
Cortex Applied Machine Learning, Twitter\\
\texttt{jo.schlemper11@imperial.ac.uk, \{jcaballero,aaitken\}@twitter.com,} \\
\texttt{joost.van.amersfoort@cs.ox.ac.uk}\\
}

\iclrfinalcopy 
\begin{document}

\maketitle
\begin{abstract}
In large scale systems, approximate nearest neighbour search is a crucial algorithm to enable efficient data retrievals. Recently, deep learning-based hashing algorithms have been proposed as a promising paradigm to enable data dependent schemes. Often their efficacy is only demonstrated on data sets with fixed, limited numbers of classes. In practical scenarios, those labels are not always available or one requires a method that can handle a higher input variability, as well as a higher granularity. To fulfil those requirements, we look at more flexible similarity measures. In this work, we present a novel, flexible, end-to-end trainable network for large-scale data hashing. Our method works by transforming the data distribution to behave as a uniform distribution on a product of spheres. The transformed data is subsequently hashed to a binary form in a way that maximises entropy of the output, (i.e. to fully utilise the available bit-rate capacity) while maintaining the correctness (i.e. close items hash to the same key in the map). We show that the method outperforms baseline approaches such as locality-sensitive hashing and product quantisation in the limited capacity regime. 
\end{abstract}

\section{Introduction}

In the modern era where we have an increasingly large amount of high-dimensional data to handle, it can be useful to have a system that can efficiently retrieve information that we care about. Examples of such systems are content-based image retieval (CBIR) \citep{datta2008image,babenko2014neural} and document/information retrieval \citep{mitra2018introduction}. In large scale systems, linear search through the dataset is prohibitive. Therefore, one often resorts to approximate methods, which allow trading off accuracy for speed. These methods are commonly called approximate nearest neighbour (ANN) methods. Another important aspect of modern systems is data locality -- ideally the data is stored a local fast disk, which restricts our representation to be quantised due to memory restrictions. Notable examples commonly used are locality-sensitive hashing (LSH) \citep{datar2004locality} and product quantisation (PQ) \citep{jegou2011product,ge2013optimized}.

Recently, deep learning has become an increasingly powerful tool for learning embeddings, due to the success of deep embedding learning \citep{oh2016deep,hermans2017defense,wu2017sampling}. These advances have motivated an approach called \emph{deep hashing} \citep{wang2016learning,erin2015deep,zhu2016deep}, where one attempts to directly obtain a hash code from an image that can be used for content-based image retrieval tasks. These methods have been shown to greatly outperform traditional approaches. However, most methods rely on explicitly incorporating the class label prediction (as opposed to constructing an affinity matrix) to improve performance, which leads to the following issues. Firstly, while exploiting the class labels can improve the discriminative capability, it makes incorporating new labels a non-trivial task. Secondly, the methods do not directly account for semantic similarities at a granular level, making it unsuitable for certain tasks such as a duplication detection. Lastly, it is common to only demonstrate the efficacy of the methods for dataset with a small number of classes ($n \leq 100$), and the generalisation for the large scale dataset seems yet to be proven. 

In this work, we propose a novel network architecture for end-to-end semantic hashing, which can be used for both deep hashing and learning an index structure \citep{kraskalearned}. Our network is inspired by a \emph{catalyser} network \citep{sablayrolles2018neural} and a supervised structured binary code (SUBIC) \citep{jain2017subic}: it explores the idea of transforming an input distribution to a uniform distribution, but directly learns to generate the hash code. Our method is also flexible such that it relies on a similarity distance, which can be neighbour ranking or class labels. We show the applicability of our model for retrieval task using publicly available data set and we experimentally show our approach outperforms baseline methods such as LSH and PQ, in particular when the available bit-rate is limited.
    
\section{Related Work}

\begin{figure}[ht!]
\begin{center}
\includegraphics[trim={0 0 0 .5cm},clip,width=1\textwidth]{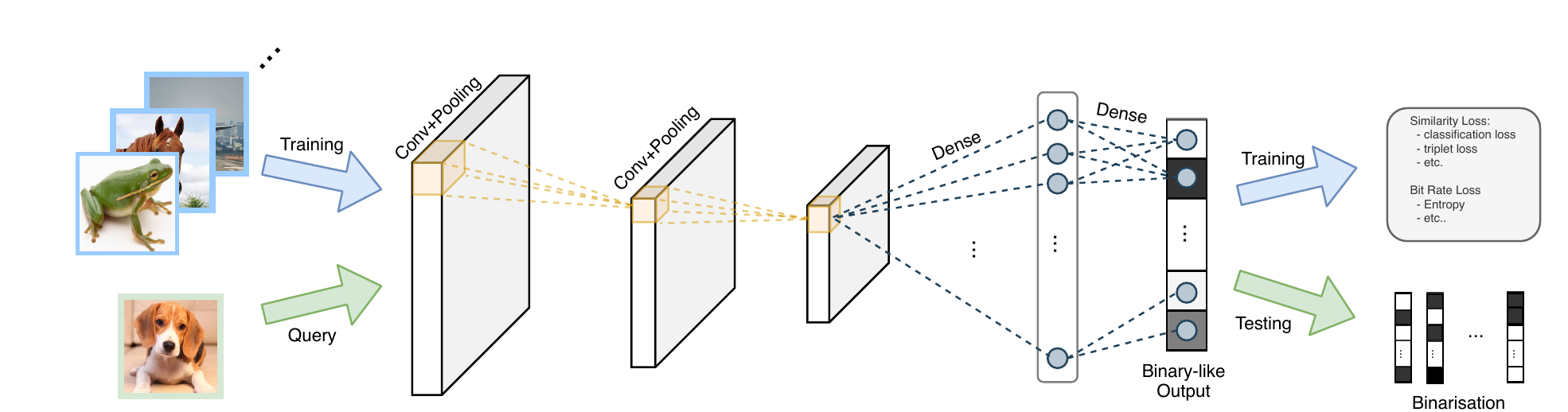}
\end{center}
\caption{Schematics of Supervised Deep Hashing.}
    \label{fig:supervised_deep_hash}
\end{figure}

In the literature of hashing, common methods include: LSH, Iterative Quantisation (ITQ) \citep{gong2013iterative} and PQ. While the first two aim to generate code in hamming space, PQ aims to represent data using a code book for a set of sub vectors. Recently deep learning-based hashing has become an active area of research \citep{lin2015deep,liu2016deep}. In particular, \emph{supervised-hashing} is a research area that is concerned with hashing and retrieving objects (e.g. text or images) belonging to specific categories. In deep-learning based hashing, often three aspects are considered for the objective function, which are:
\begin{enumerate}
    \item How to preserve semantic similarities of the inputs in their generated hash codes?
    \item How to devise a continuous representation that can be trained using a neural network, which simultaneously minimises the discrepancy from test-time discretisation/binarisation?
    \item How can we optimise the available bit-rate of a database (the output space), which can help minimise the collision probability?
\end{enumerate}

The first challenge is often addressed by using a metric learning approach, such as contrastive loss \citep{hadsell2006dimensionality}, triplet loss \citep{schroff2015facenet, hermans2017defense} and their $n$-way extensions \citep{chen2017beyond}. The main drawback of these losses is that it is difficult to optimise them in a high dimensional space due to the curse of dimensionality \citep{friedman2001elements}. Therefore, whenclass labels are available, it can be more effective to utilise those \citep{jain2017subic}. Doing so induces a dependence on quality and availability of the class labels. In fact, as pointed out in \cite{sablayrolles2017should}, retrieving objects based on classification puts an upper bound on the performance for the recall value. It can be more desirable to optimise the model on a more flexible objective which allows granular hashing based on semantic similarity, rather than solely the label information. The second challenge is a product of the non-differentiability of a naive discretisation step. Therefore at train time, one can resort to non-linearities such as $sigmoid$ and $tanh$, or continuous functions with better properties \citep{cao2017hashnet, cao2018deep}. The last point is usually handled by variants of entropy-based regularisation \citep{jain2017subic}. 

We also provide a more in-depth account of SUBIC \citep{jain2017subic} and catalyser network \citep{sablayrolles2018neural} as these networks form a foundation of our network architecture. 

\paragraph{SUBIC} Let $x \in \mathbb{R}^{N}$ be an input data (e.g. an image) and $y \in \mathcal{Y} = \{ 1, \dots, C \}$ is the the class label. Given an image $z$, SUBIC outputs a structured binary code $b$, which is expressed as $M$ $K$-blocks: $b = [b^{(1)}; \dots; b^{(M)}]$ with $b^{(i)} \in \mathcal{K}_K = \{ d \in \{0,1\}^K \text{ s.t. } \|d\|_1 = 1 \}$. This is achieved by the following network: 

\begin{align}
    b = f_{\text{SUBIC}}(x)  = \sigma\circ f_{\text{enc}} \circ f_{\text{feat}}(x)
\end{align}

where $f_{\text{feat}}:\mathbb{R}^N\to \mathbb{R}^{N'}$ is a feature extractor, $f_{\text{enc}}:\mathbb{R}^{N'}\to\mathbb{R}^{MK}$ is a hash encoder, $\sigma$ is a nonlinearity which applies $K$-softmax function to each of the $M$ block and ``$\circ$`` is a composition operator. During training, the $K$-blocks are relaxed into $K$-simplicies: $\widetilde{b} = [\widetilde{b}^{(1)}; \dots; \widetilde{b}^{(M)}]$ with $\widetilde{b}^{(i)} \in \red{\Delta}_K = \{d \in \red{[}0,1\red{]}^K \text{ s.t. } \|d\|_1 = 1\}$.  The novelty of SUBIC is to fit a classification layer $f_{\text{clf}}:\mathbb{R}^{MK}\to\mathbb{R}^{C}$to learn a discriminative binary code.  Given minibatch $\mathcal{B} = \{(x_i,y_i)\}_{i=1}^B $, the network is trained by minimising the following loss function:

\begin{align}
\label{eq:subic_loss}
\mathcal{L}(\mathcal{B}) = \underbrace{ \frac{1}{B} \sum_{i=1}^B  \log s[y_i]}_{\text{Cross Entropy}} 
+ 
\gamma \underbrace{\frac{1}{B} \sum_{i=1}^B E(\widetilde{b}_i )}_{\text{Entropy}}
- \mu \underbrace{ E \left(\frac{1}{B} \sum_{i=1}^{B} \widetilde{b}_i \right)}_{\text{ Batch Entropy}}
\end{align}

where $s = f_\text{clf}(\widetilde{b})$, $\widetilde{b} = f_{\text{SUBIC}}(x) $ and $E$ is mean entropy of $K$ blocks:

\begin{align}
    E(b) = - \frac{1}{M} \sum_{m=1}^M \sum_{k=1}^K b^{(m)}[k] \log_2 b^{(m)}[k]
\end{align}

The idea of the \emph{Entropy} term is to encourage the network output to become one-hot like. On the other hand, the \emph{Negative Batch Entropy} term encourages the uniform block support so that the available bit rate is fully exploited. At test time, $\sigma$ is replaced by a block-wise $\arg\max$ operation, where the binary code is obtained by setting the maximum activated entry in each block to 1, the rest to 0.

\paragraph{Catalyser network}

Let $x$ be the input, $z=f_\text{cat}(x)$ be the network embedding before quantisation is applied. Let $b = f_\text{quant}(z)$ be the quantised representation. The idea of the catalyser network is to embed data uniformly on an $\ell_2$-sphere, i.e. $z \in S^{N}$, which is subsequently encoded by an efficient lattice quantiser. The network is trained by minimising teh \emph{triplet rank} loss \citep{hermans2017defense} and maximising the entropy loss. Given a triplet of input $(x_a, x_p, x_n)$ (anchor point, positive sample and negative sample respectively), the loss is defined as:

\begin{align}
    \mathcal{L}_\text{tri} = \left[ \|f_{cat}(x_a) - f_{cat}(x_p) \|_2 - \| f_{cat}(x_a) - f_{cat}(x_n) \|_2 + \alpha \right]_+
    \label{eq:triplet}
\end{align}

for margin $\alpha$. For entropy regularisation, Kozachenko and Leonenko (KoLeo) entropy estimator is used as a surrogate function:

\begin{align}
H_n = \frac{\alpha}{n} \sum_i^n \log ( \rho_{n,i} ) + \beta, \quad \rho_{n,i} = \min_{i \neq j} \| f_{cat}(x_i) - f_{cat}(x_j) \|
\end{align}

The equation is simplified to:

\begin{align}
    \mathcal{L}_{KoLeo} = \sum_i^n \log(\rho_{n,i})
    \label{eq:koleo}
\end{align}

The geometric idea is to ensure any two points are sufficiently far from each other, where the penalty decays logarithmically. The network then quantises the output using a Gosset Code, we refer the interested reader to \cite{sablayrolles2018neural} for more details.

\begin{figure}[ht!]
\begin{center}
\includegraphics[trim={0 1.5cm 0 .1cm},clip,width=1\textwidth]{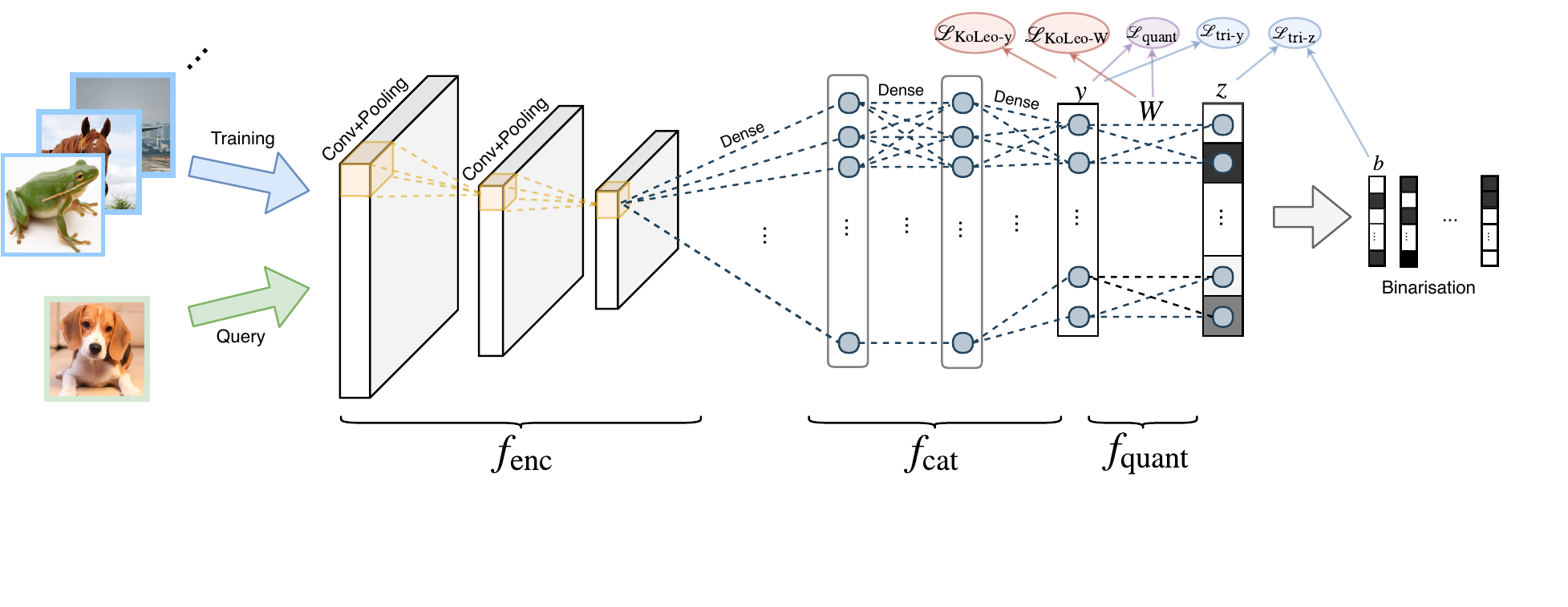}
\end{center}
\caption{Schematic of the proposed architecture.}
    \label{fig:architecture}
\end{figure}

\section{Proposed Approach}

While SUBIC is effective, its application is limited to cases where classification labels are available. On the other hand, catalyser networks are more flexible, but not end-to-end trainable. In this work, we propose a flexible approach which incorporates the benefit of both and mitigates their limitations.

The proposed network is composed of three components: a feature extractor $f_\text{enc}$, a catalyser $f_\text{cat}$ , and a quantiser $f_\text{quant}$, see Fig. \ref{fig:architecture}. Given an input image $x$, the network directly generates a hash, which is a structured binary code as in SUBIC: $b \in [0,1]^M_K$.  Let $y = f_\text{cat} \circ f_\text{enc} (x)$, $b = f_\text{quant}(y)$. The difference between catalyser networks and our work is that we learn the quantisation network $f_\text{quant}$, making the architecture end-to-end trainable. Our quantiser $f_\text{quant}(y)$ is given by $z^{(m)} = \sigma_k(W y)$, where $\sigma_k$ is a block-wise K-softmax. 

\begin{figure}[ht]
\begin{center}
\includegraphics[width=0.5\textwidth]{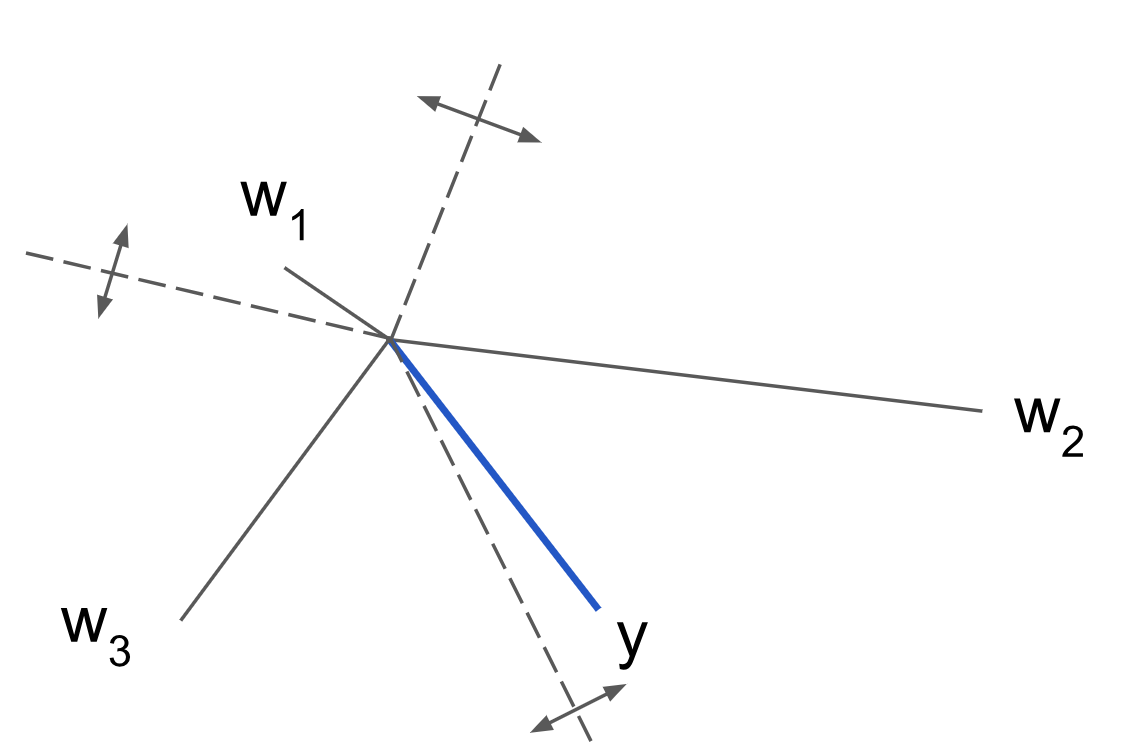}
\end{center}
\caption{The geometry of the linear layer. The dashed line defines the assignment boundary for each $w_i$;s}
    \label{fig:intuition}
\end{figure}

For simplicity, consider each $K$-block separately. Our key insight is the following: a fully connected layer is simply a dot product between $y$ and the row vectors of $W$. We have $z^{(m)} = \sigma_k(\langle w_1, y \rangle, \dots, \langle w_K, y \rangle)$. Since at test-time, binarisation is done by selecting the maximally activated entry (within each $K$-block), this is equivalent to selecting the row vector with the smallest angular difference. This can be visualised using row vectors, $w_1, \dots, w_k$ which linearly partition the output space, and the decision boundary is extending from the origin (Fig. \ref{fig:intuition}). 

Let $p(y)$ denote the probability distribution of the catalyser output taking a specific value in $y \in S^{K-1} $. To achieve the maximal entropy of $f_\text{quant}(y)$ (i.e. to maximise the used bit-rate), one can assign $y$ to each row vector $w_i$'s with an equal probability. Geometrically, this can be seen as equally partitioning the support of $p(y)$ by $w_i$'s, where $\text{supp}(p(y)) = \{y \in S^{K-1} \hspace{0.1em} |\hspace{0.1em}  p(y  = f_{cat} \circ f_{enc} (x)) > 0  \}$. If $y$ is \emph{uniformly} distributed on $S^{K-1}$, then it is sufficient to uniformly distribute $w_i$'s to partition the $S^{K-1}$ equally. This gives us the following strategy: we (1) encourage the distribution of $p(y)$ to be uniform on a sphere and (2) uniformly distribute $w_i$'s on the sphere. We can achieve both by using KoLeo entropy estimators in Eq. \ref{eq:koleo}:

\begin{align}
\label{eq:koleo_y}
    \mathcal{L}_\text{KoLeo-y} & = \sum_i^n \log(\rho_{n,i}), \quad \rho_{n,i} = \min_{j \neq i} \| y_i - y_j \| \\
\label{eq:koleo_w}
    \mathcal{L}_\text{KoLeo-W} & = \sum_i^n \log(\tau_{n,i}), \quad \tau_{n,i} = \min_{j \neq i} \| w_i - w_j \|
\end{align}

However, it is likely that a perfect uniform distribution cannot be achieved by the training, especially once combined with other deep embedding losses. To circumvent this, we add the following:

\begin{align}
\label{eq:loss_quant}
    \mathcal{L}_\text{quant} & = \sum_i \| y_i - w_{\iota(y_i)} \|
\end{align}

where $j = \iota(y_i) = \arg\min_{j'} \| y_i - w_{j'} \|$ (i.e. the index of the closest $w_j$ to $y$). By minimising Eq. \ref{eq:loss_quant}, the row vectors $w_i$'s will be gravitated towards the probability mass of $p(y)$.

The remaining aspect is similar to the previous approaches: we minimise triplet loss to ensure similar points are embedded closely. For this, we can minimise triplet loss either in the output space of catalyser $\underbrace{S^{K-1} \times \dots \times S^{K-1}}_{M}$ or the relaxed output space $\Delta_K^M$. Indeed, it is beneficial to directly optimise in the final target space. However, interestingly, it turns out that minimising triplet rank loss in simplex is difficult due to the fact that most points are very close to each other in high dimensional simplicies (see Appendix), yielding training instability. We mitigate this issue by using \emph{asymmetrical} triplet loss:

\begin{align}
\mathcal{L}_\text{tri-z} = \left[ \|z_a - b_p \| - \| z_a - b_n \| + \alpha \right]_+
    \label{eq:modified_triplet}
\end{align}

where $z_a$, $z_p$, $z_n$ are the anchor, positive and negative points in the embedded space respectively, $\alpha$ is a margin, $b_*$'s are the discretised point (i.e. by replacing softmax by argmax). Note that sampling $b_*$'s is non-differentiable due to argmax, but we can nevertheless backpropagate the information using straight-through estimator (STE) proposed by \cite{DBLP:journals/corr/BengioLC13}, which has resemblance to stochastic graph computation approaches \citep{maddison2016concrete}. Secondly, the loss becomes zero if $b_p$ and $b_n$ share the same binary representation. We empirically found incorporating the triplet loss in $\underbrace{S^{K-1} \times \dots \times S^{K-1}}_{M}$ was a useful additional loss to overcome this issue. The final objective is thus:

\begin{align}
\label{eq:objective}
    \mathcal{L} = \mathcal{L}_\text{tri-z} + \lambda_1 \mathcal{L}_\text{tri-y} + \lambda_2 \mathcal{L}_\text{KoLeo-y} + \lambda_3 \mathcal{L}_\text{KoLeo-W} + \lambda_4 \mathcal{L}_\text{quant}
\end{align}

where $\lambda_i$'s are hyper-parameter to be optimised. Note that for triplet loss,  $\ell_2$-normalising $y$ and the rows of $W$ is important as otherwise arbitrary scaling can make the training unstable. Secondly, to reduce the parameters, we partition $W$ to only take each $K$-blocks and learn $W_1, \dots,  W_k$, where $W_i \in \mathbb{R}^{K \times K}$.\footnote{Ideally, $M$ $K$-blocks are decorrelated to remove the redundancy. This is left as a future work.} Finally, note that feature extractor and catalyser are only optimised with respect to $\mathcal{L}_\text{tri-z}, \mathcal{L}_\text{tri-y}$, and $\mathcal{L}_\text{KoLeo-y}$, whereas the quantiser weight $W$ is optimised only with respect to $\mathcal{L}_\text{tri-z}$, $\mathcal{L}_\text{KoLeo-w}$ and $\mathcal{L}_\text{quant}$. In particular, Eq. \ref{eq:loss_quant} is  only minimised by $W$.

\subsection{Encoding and distance computation}

Given a set of data points, we encode via $b = f_\text{quant} \circ f_\text{cat} \circ f_\text{feat}(x)$. The resulting vector can be compressed by storing indices of one-of-$K$ vectors, which only requires $M\log(K)$ bits. The distance between two compressed points can be given by Euclidean distance: $\| b_i - b_j \|_2$, which can be efficiently computed by \emph{$M$ look up}: $\sum_i^M b^{(i)}_\text{query}[i(b^{(i)})]$, where $i(.)$ is the index of $b^{(i)}$ having one. One can also perform \emph{asymmetric distance comparison} (ADC), which in case $b_\text{query}$ is replaced by $y_\text{query} = f_\text{cat} \circ f_\text{feat}(x)$, the data representation prior to quantisation.  

\subsection{Network Implementation} 

For the feature encoder, a pre-trained network can be used, such as VGG or Resnet architectures. The catalyser was implemented using a fully connected network with 2 hidden layers, each having 256 features, and a final layer which maps the dimension to $d=MK$. We used batch-normalisation and Rectified Linear Unit (ReLU) for non-linearity, on all layers except the final one. The quantiser are $M$ separate fully connected layers with $K$ features. The overall network was trained using Adam with $\alpha=10^{-3},\beta_1 = 0.9, \beta_2 = 0.999$. The convergence speed of the network depends on the size of $K$, but usually sufficient performance can be obtained within 3 hours of training. 

\begin{figure}[ht]
\begin{center}
\includegraphics[width=0.6\textwidth]{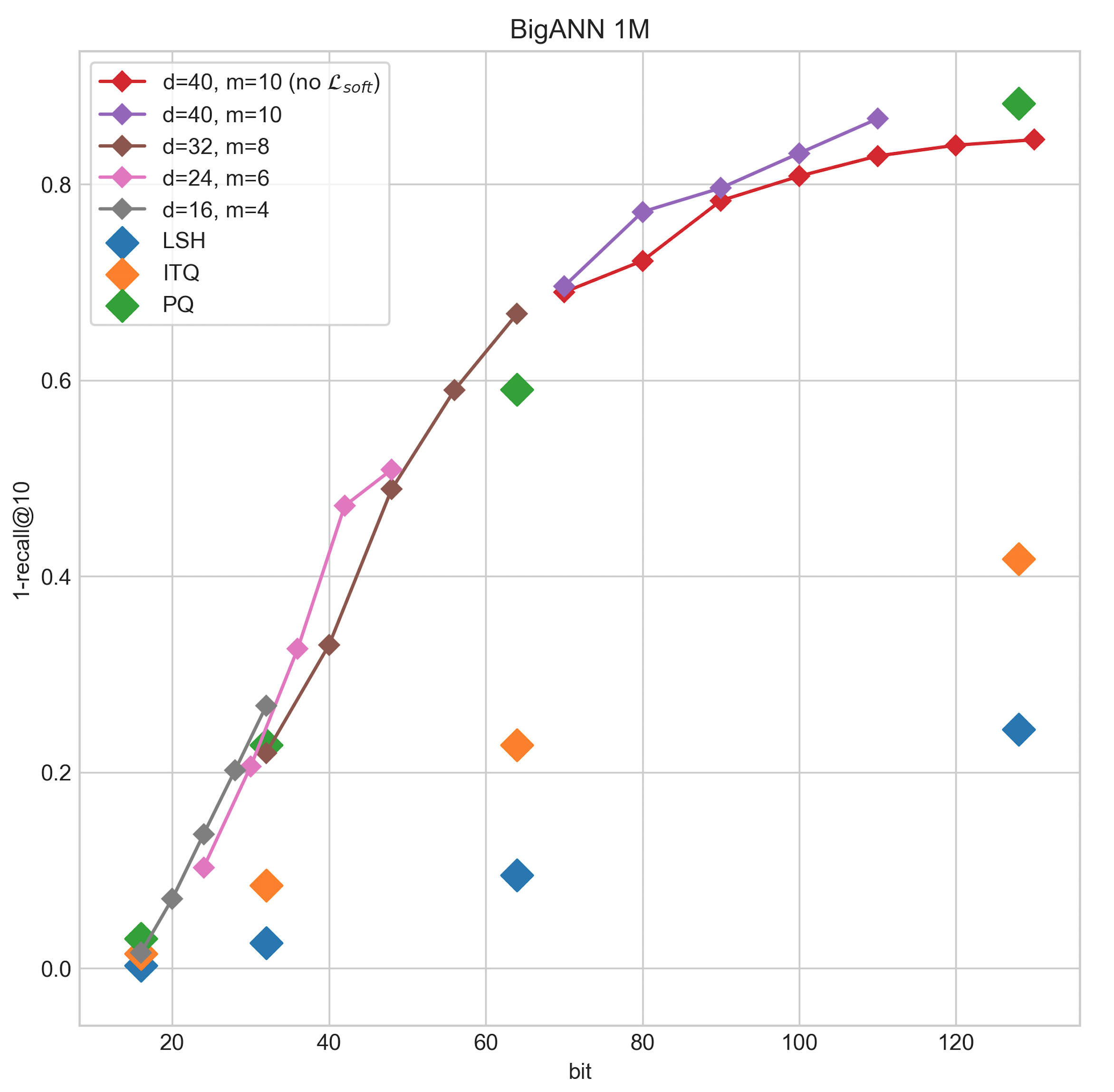}
\end{center}
\caption{1-Recall@10 for BigANN1M dataset.}
    \label{fig:bigann_result}
\end{figure}

\section{Experiment}

\subsection{BigANN1M}

We evaluate our proposed approach using BigANN1M dataset\footnote{Publicly available at http://corpus-texmex.irisa.fr/}: the dataset contains a collection of 128 dimensional SIFT feature vectors. As the input is already feature vectors, we set $f_\text{enc} = \text{id}$. Training data contains 30,000 points, test data contains 10,000 query points and 1 million database points. For each point, we labelled the top $k=10$ nearest points in terms of Euclidean distance to be the neighbours for triplet loss.

For evaluation, we used the metric 1-Recall@K=10, which measures the probability of retrieving the true first neighbour within the first 10 candidates. We compare to LSH, ITQ and PQ for the baseline methods. For PQ, we chose $K=256$ for each sub-vector and varied the values of $M$ to achieve the desired bit-length $B \in \{16, 32, 64, 128\}$. 

The result is summarised in Fig. \ref{fig:bigann_result}. For the proposed method, we varied the number of $d$, $M$ and $K$ to get different number of bits. One can see that the performance of the proposed approach is comparable to PQ, but better for lower number of bits. Note that ITQ and LSH uses symmetric distance comparison so it is an unfair comparison. We also compared the proposed model with and without $\mathcal{L}_{quant}$ and we see a noticeable improvement. We speculate that this is because, while even without the loss, since the points are uniformly distributed it can achieve sufficient level of reconstruction, by minimising the quantisation loss, we remove the ``gaps`` in $\text{supp}(p(y))$.








\subsubsection{Visualisation}

\begin{figure}[ht]
\begin{center}
\includegraphics[width=\textwidth]{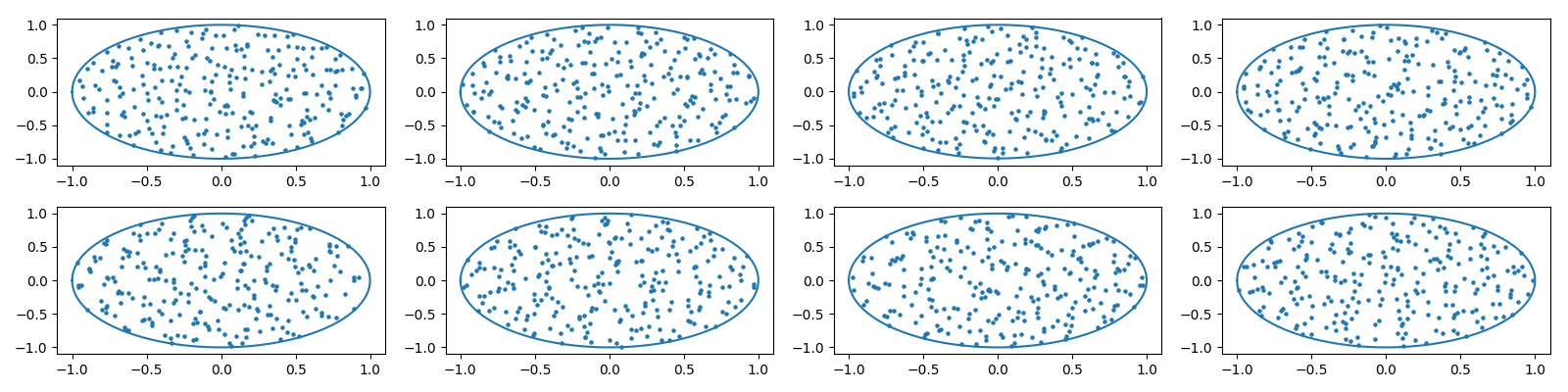}
\end{center}
\caption{The distribution of the weights $w_i$ of the quantiser without $\mathcal{L}_\text{quant}$. Each sphere corresponds to one sub-block. For each sub-block, we randomly select 2 axes for visualization.}
    \label{fig:q_w}
\end{figure}

We visualise the learnt weight vectors of the quantiser $f_\text{quant}$. For each $M$ sub-block, we randomly select 500 row vectors. Then we visualise 2 axes of these vectors (i.e. a projection onto 2 dimensional plane rather than using dimensionality reduction techniques).  Without using $\mathcal{L}_\text{quant}$, the weights are uniformly distributed on $k$-sphere (Fig \ref{fig:q_w}). However, when the loss is introduced, we see the mass of the rows concentrates on a more local area. 

\begin{figure}[ht]
\begin{center}
\includegraphics[width=\textwidth]{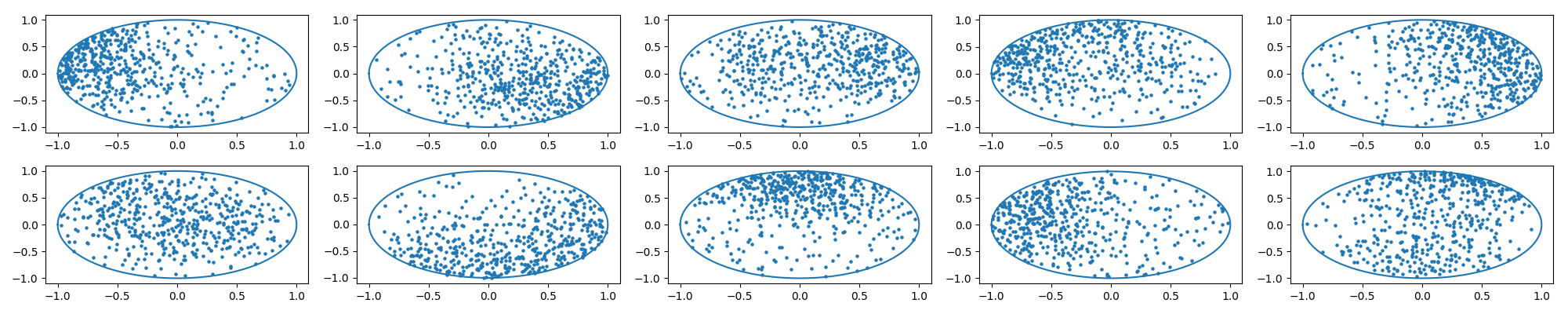}
\end{center}
\caption{The distribution of weights $w_i$ of the quantiser with $\mathcal{L}_\text{quant}$. Each sphere corresponds to one sub-block. For each sub-block, we randomly select 2 axes for visualization.}
    \label{fig:q_w_l}
\end{figure}

\section{Conclusion}

In this work, we proposed a deep neural network which can perform end-to-end hashing of input, which only requires the knowledge of similarity graph, which is a slightly more relaxed constraint than class labels. The network operates by transforming the input space into a uniform distribution by penalising the cost given by KoLeo differential estimator, which was quantised by weight vectors uniformly distributed in its support. The network performs comparatively to the baseline methods, however, there is plenty of room for improvement. In the future, it will be interesting to impose a different prior on the distribution of the simplex, e.g. via Dirichlet distribution, to help control the output distribution, rather than relying on a uniform distribution.

\subsubsection*{Acknowledgements}

We thank Lucas Theis, Ferenc Husz\'{a}r, Hanchen Xiong and Twitter London CAML team for their valuable insights and comments for this work. 

\bibliography{iclr2019_conference}
\bibliographystyle{iclr2019_conference}

\newpage

\section{Appendix}

\subsection{Distribution of pairwise distances on surfaces in n dimension}
In the main manuscript, we argued that it is difficult to directly train triplet rank loss on high-dimensional simplex. Here we show how points on $n$-dimensional objects are distributed in high dimension as a part of the argument. 

\begin{figure}[ht]
\begin{tabular}{ccc}
 \includegraphics[width=0.3\textwidth]{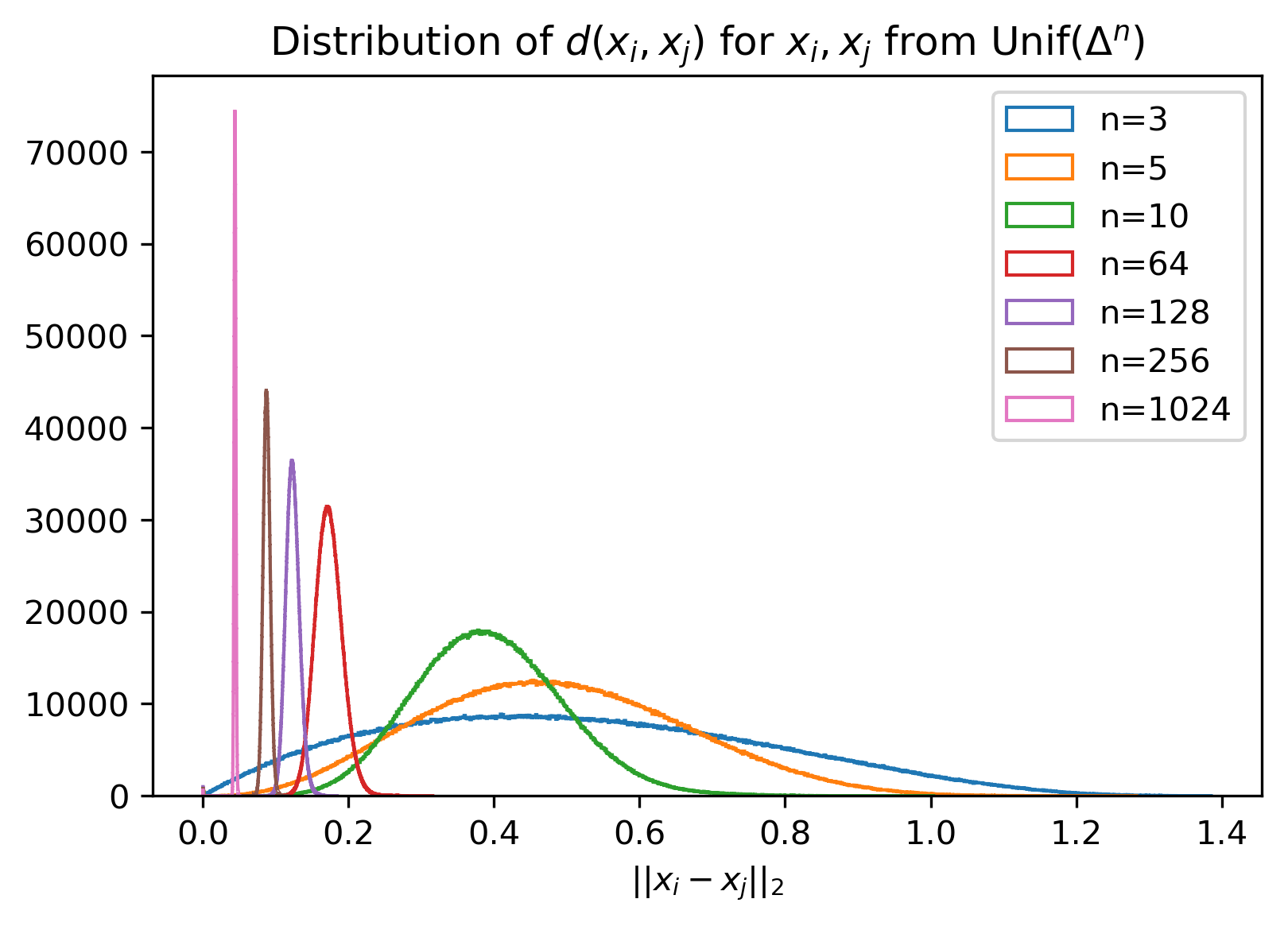} 
 & \includegraphics[width=0.3\textwidth]{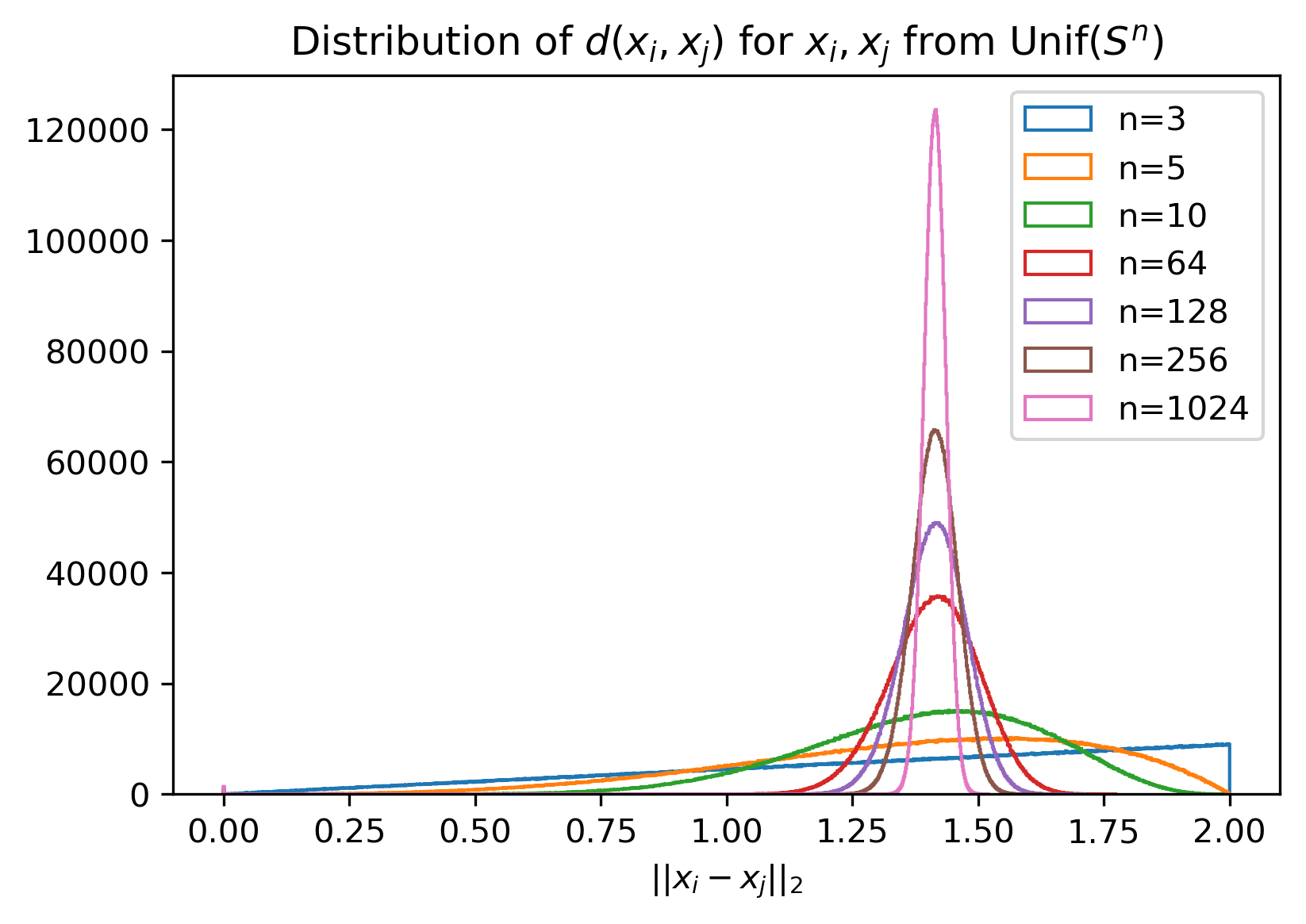}
 & \includegraphics[width=0.3\textwidth]{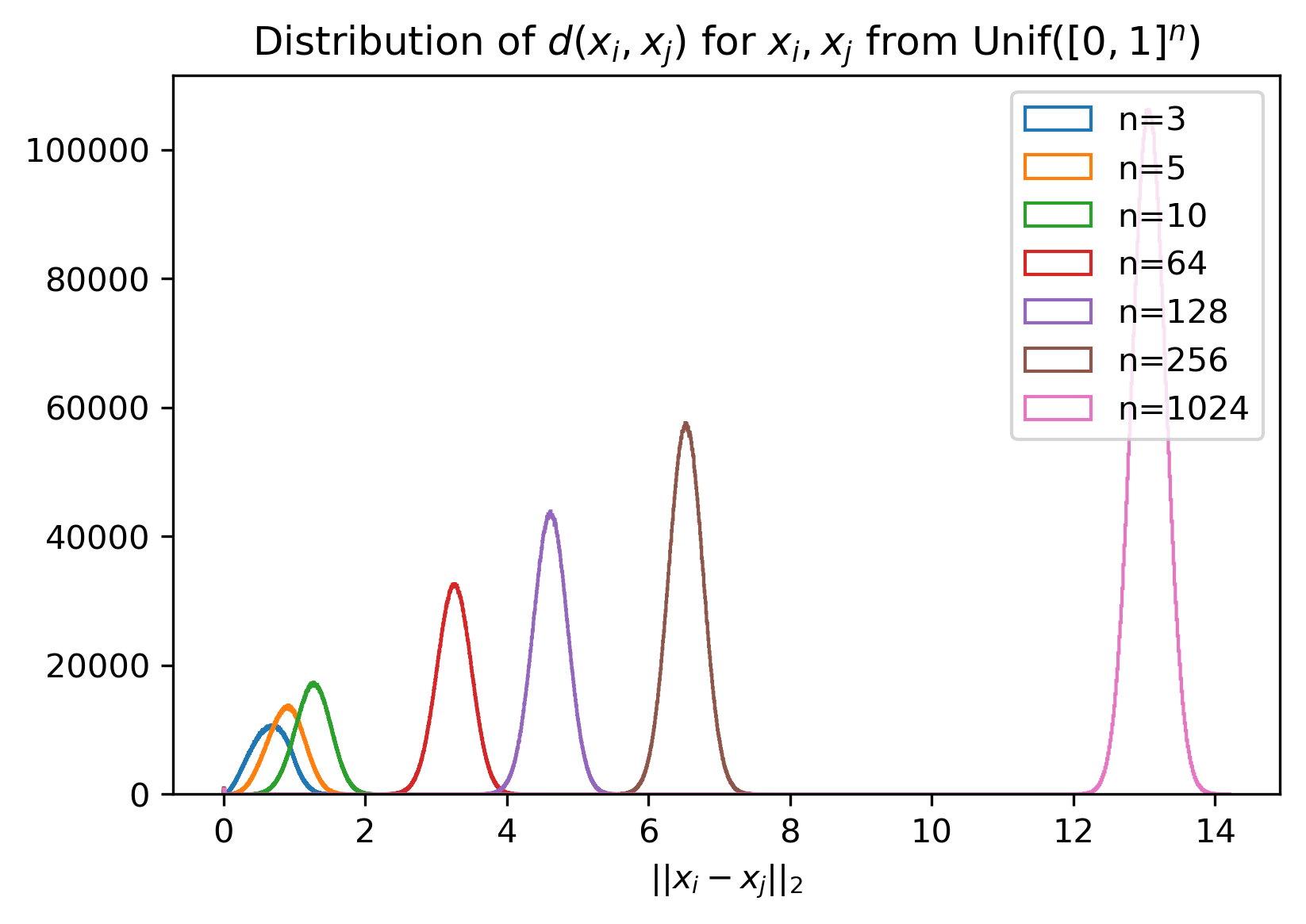} \\
\end{tabular}
\caption{The distribution of $\ell_2$ distance between two random points on $n$-dimensional objects}
\label{fig:dist_pts}
\end{figure}

\subsubsection{Interior of Simplex}

\begin{figure}[ht]
\begin{center}
\includegraphics[width=\textwidth]{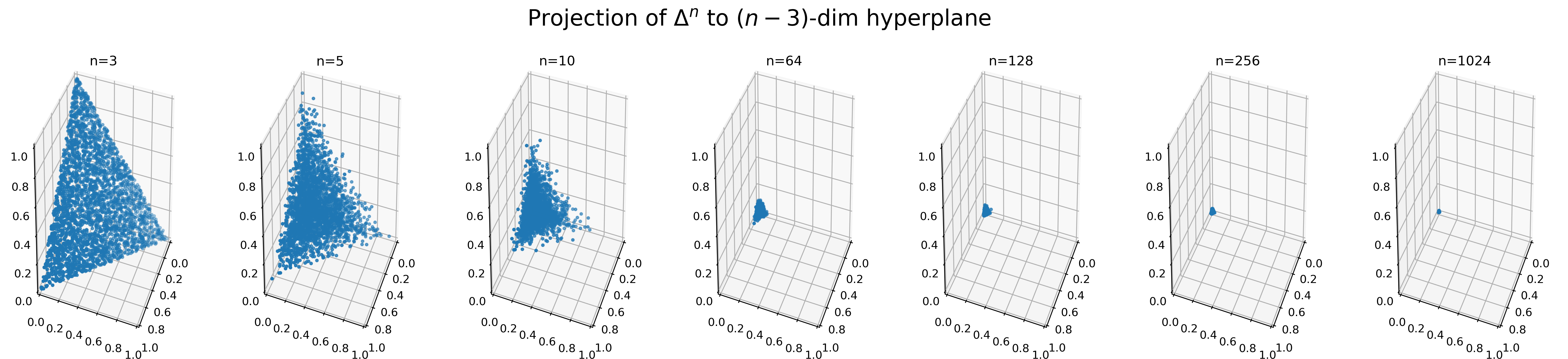}
\end{center}
\caption{3D projection of high-dimensional simplex}
    \label{fig:simplex}
\end{figure}

We use Dirichlet distribution with concentration parameter $a_1 = \dots = a_n = 1$ to sample points uniformly in the interior of $n$-dimensional simplex. As one can see from Fig \ref{fig:simplex}, as the dimension increases, the the points become more concentrated around the center of simplex $\mu=(1/n,\dots, 1/n)$. The distribution of $\ell_2$ distance between two uniformly sampled points on $n$-simplex also sharply concentrates around small value, as it can be seen in \ref{fig:dist_pts}. However, in the case of triplet rank loss, we would like to guarantee sufficiently high margin to ensure the separation between different classes. For example, the distance from any of the vertices to the centre of simplex is $\frac{\sqrt{n-1}}{\sqrt{n}}$ and the distance between two vertices is $\sqrt{2})$. We empirically saw that often the network collapses to predicting just $\mu$ and it is difficult to satisfy meaningful margin $\alpha$ as well as pushing the points approach towards one of the vertices.  

\subsubsection{Surface of n-Sphere}
We sample points uniformly on $n$-sphere by first sampling $n$ from $\mathcal{N}(0,1)$, followed by $\ell_2$-normalisation. In this case, the distribution of distances between two uniformly sampled points on $n$-sphere is given by: $p(d) \propto d^{n-2}[1 - \frac{1}{4}d^2]^{\frac{n-3}{2}}$ \citep{wu2017sampling}. As $n\to\infty$, the probability distribution converges to $\mathcal{N}(\sqrt{2}, \frac{1}{2n})$. In this case, there is sufficient space left between majority of points, which is why we speculate that it is easier to train with triplet rank loss. Note that this however also means that since all points are already $\sqrt{2}$ far, careful \emph{negative example mining} becomes very important to yield useful gradient. 

\begin{figure}[ht]
\begin{center}
\includegraphics[width=\textwidth]{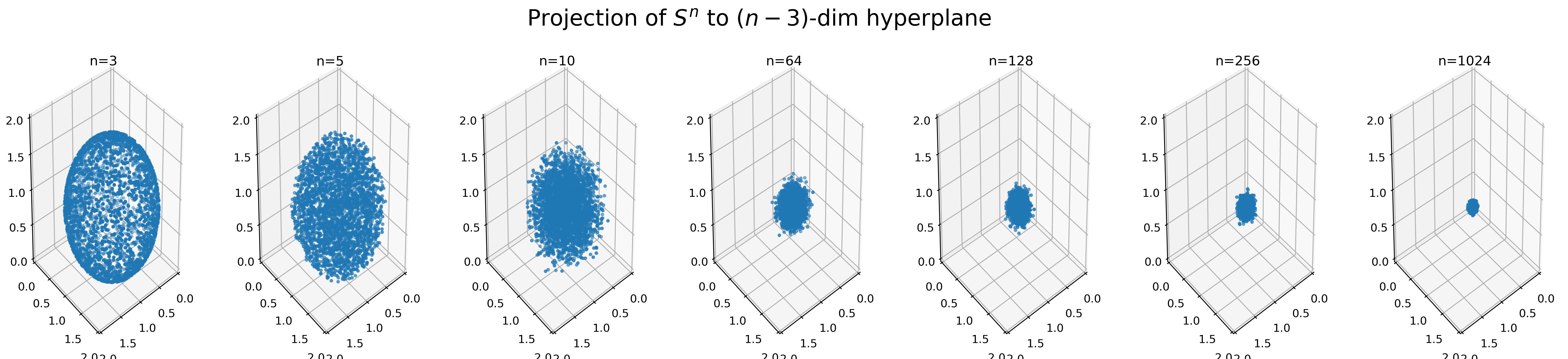}
\end{center}
\caption{3D projection of high-dimensional $n$-sphere}
    \label{fig:sphere}
\end{figure}

\subsubsection{Interior of n-cube}

We also study the distribution of distances between two random points in the interior of a hypercube. Here, the distances gets increasingly large as $n\to \infty$. Therefore, hypercube would have been an alternative shape we could use as a domain for hashing, which could be interesting for future work. However, In this case, we could use sigmoid function to set the range, but this could result in gradient saturation. 

\begin{figure}[h]
\begin{center}
\includegraphics[width=\textwidth]{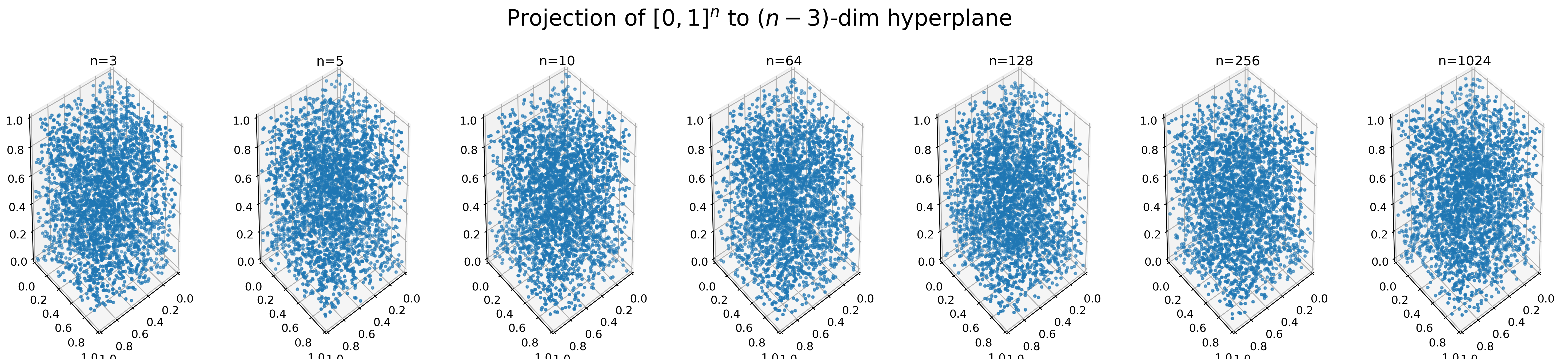}
\end{center}
\caption{3D projection of high-dimensional $n$-cube}
    \label{fig:cube}
\end{figure}

\end{document}